\documentclass[10pt,twocolumn,letterpaper]{article}

\usepackage[pagenumbers]{cvpr} % To force page numbers, e.g. for an arXiv version

\usepackage{graphicx}
\usepackage{amsmath}
\usepackage{amssymb}
\usepackage{booktabs}
\usepackage{verbatim}

\usepackage{bbm}
\usepackage{hyperref}
\usepackage{url}
\usepackage{caption}
\usepackage{xspace}
\usepackage{array}
\usepackage{makecell}
\usepackage[ruled]{algorithm2e}
\usepackage{algpseudocode}
\algnewcommand{\algorithmicand}{\textbf{ and }}
\algnewcommand{\algorithmicor}{\textbf{ or }}
\algnewcommand{\OR}{\algorithmicor}
\algnewcommand{\AND}{\algorithmicand}

\usepackage{enumitem}
\usepackage{tabu}
\usepackage{multirow}
\usepackage{xcolor}
\usepackage{makecell}
\usepackage{xcolor,colortbl}
\usepackage{dictsym}
\definecolor{Gray}{gray}{0.90}
\definecolor{citecolor}{HTML}{0071bc}
\hypersetup{
    colorlinks=true,
    linkcolor=red,
    filecolor=magenta,      
    urlcolor=magenta,
    citecolor=citecolor,
}
\usepackage{pifont}% http://ctan.org/pkg/pifont
\newcommand{\xmark}{\ding{55}}%

\newcommand{\name}{BoxTeacher}

\newcommand{\apm}{AP}
\newcommand{\aps}{AP & AP$_{50}$ & AP$_{75}$}
\newcommand{\apl}{AP$_{50}$ & AP$_{75}$ & AP$_{s}$ & AP$_{m}$ & AP$_{l}$}
\newcommand{\tline}{\Xhline{1pt}}

\definecolor{mygray}{gray}{0.93}
\newcommand{\gt}[1]{\textcolor{gray}{#1}}

\makeatletter\renewcommand\paragraph{\@startsection{paragraph}{4}{\z@}
  {.5em \@plus1ex \@minus.2ex}{-.5em}{\normalfont\normalsize\bfseries}}\makeatother

\usepackage[capitalize]{cleveref}
\crefname{section}{Sec.}{Secs.}
\Crefname{section}{Section}{Sections}
\Crefname{table}{Table}{Tables}
\crefname{table}{Tab.}{Tabs.}

\def\cvprPaperID{9371} % *** Enter the CVPR Paper ID here
\def\confName{CVPR}
\def\confYear{2023}

\begin{document}

%%%%%%%%% TITLE - PLEASE UPDATE
\title{BoxTeacher: Exploring High-Quality Pseudo Labels for Weakly Supervised Instance Segmentation}

\author{
Tianheng Cheng$^{1,\star}$,
Xinggang Wang$^{1,\dagger}$,
Shaoyu Chen$^{1,\star}$,
Qian Zhang$^2$,
Wenyu Liu$^{1}$\\
[1mm]
$^1$~School of EIC, Huazhong University of Science \& Technology\\ 
$^2$~Horizon Robotics
\\ 
\normalsize{
\url{https://github.com/hustvl/BoxTeacher}}
}

\maketitle

\let\thefootnote\relax\footnotetext{$^\star$ This work was done when Tianheng Cheng and Shaoyu Chen were interns at Horizon Robotics. $^\dagger$  Xinggang Wang is the corresponding author: \texttt{xgwang@hust.edu.cn}}

\begin{abstract}
Labeling objects with pixel-wise segmentation requires a huge amount of human labor compared to bounding boxes.
Most existing methods for weakly supervised instance segmentation focus on designing heuristic losses with priors from bounding boxes.
While, we find that box-supervised methods can produce some fine segmentation masks and we wonder whether the detectors could learn from these fine masks while ignoring low-quality masks.
To answer this question, we present \name{}, an efficient and end-to-end training framework for high-performance weakly supervised instance segmentation, which leverages a sophisticated teacher to generate high-quality masks as pseudo labels.
Considering the massive noisy masks hurt the training, we present a mask-aware confidence score to estimate the quality of pseudo masks, and propose the noise-aware pixel loss and noise-reduced affinity loss to adaptively optimize the student with pseudo masks.
% The student can adaptively learn from the pseudo labels with pixel-wise and affinity loss.
Extensive experiments can demonstrate effectiveness of the proposed \name{}.
Without bells and whistles, \name{} remarkably achieves $35.0$ mask AP and $36.5$ mask AP with ResNet-50 and ResNet-101 respectively on the challenging COCO dataset, which outperforms the previous state-of-the-art methods by a significant margin and bridges the gap between box-supervised and mask-supervised methods.
% The code and models will be available later.
\end{abstract}

\section{Introduction}

Instance segmentation, aiming at recognizing and segmenting objects in images, is a fairly challenging task in computer vision.
Fortunately, the rapid development of object detection methods~\cite{fasterrcnn,FCOSTianSCH19,DETRCarionMSUKZ20} has greatly advanced the emergence of numbers of successful methods~\cite{HeGDG17,CascdeCaiV21,SOLOWangKSJL20,SOLOV2WangZKLS20,TianSC20,YolactBolyaZXL19} for effective and efficient instance segmentation.
With the fine-grained human annotations, recent instance segmentation methods can achieve impressive results on challenging the COCO dataset~\cite{COCOLinMBHPRDZ14}. 
Nevertheless, labeling instance-level segmentation is much complicated and time-consuming, \eg, labeling an object with polygon-based masks requires $10.3\times$ more time than that with a 4-point bounding box~\cite{pointsup}.
\begin{figure}
\centering
\includegraphics[width=1.0\linewidth]{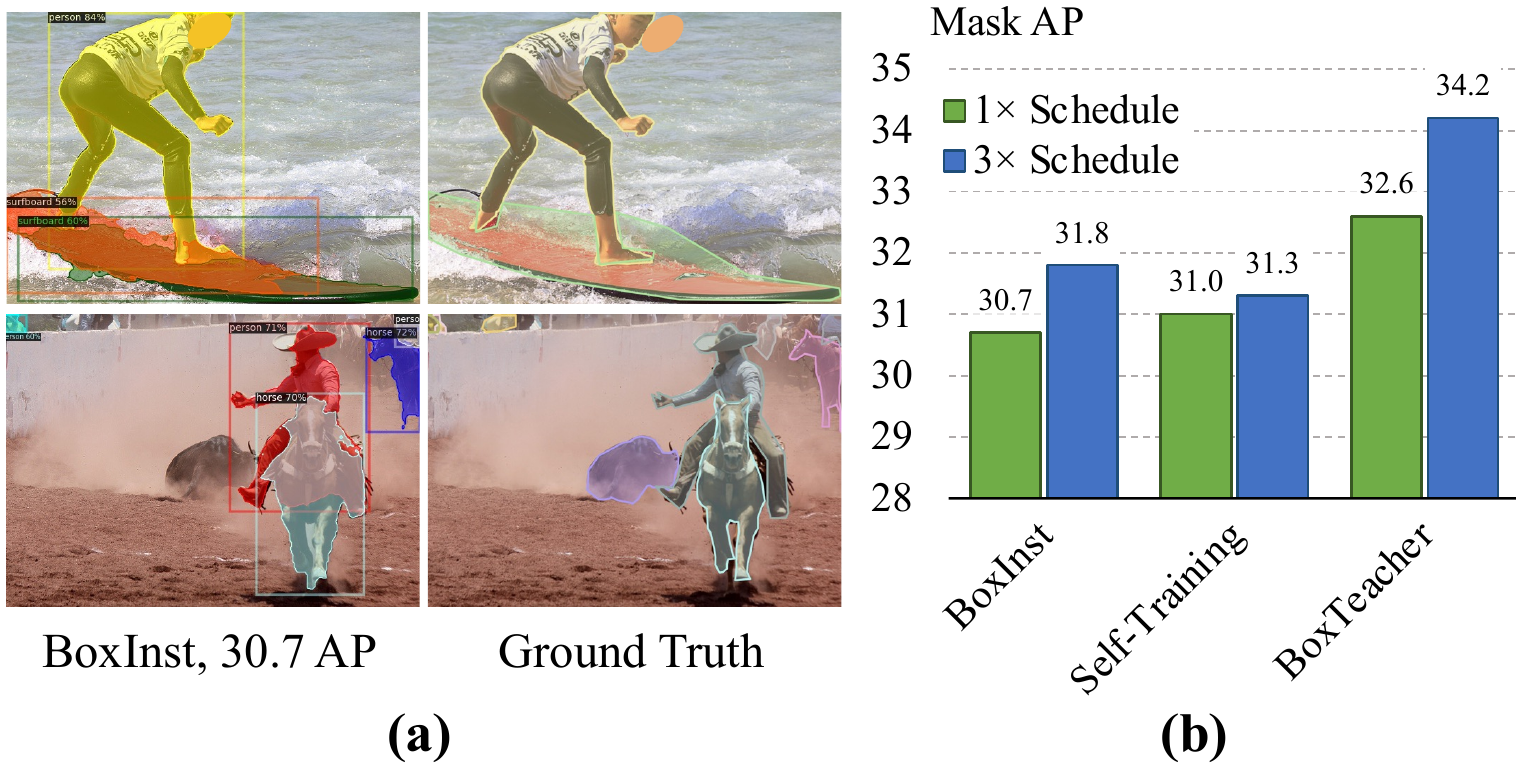}
\vspace{-18pt}
\caption{\textbf{(a) Segmentation Masks from BoxInst.} BoxInst (ResNet-50~\cite{HeZRS16}) can produce some fine segmentation masks with weak supervisions from bounding boxes and images. \textbf{(b) Self-Training with Pseudo Masks on COCO \texttt{val}.} We explore the self-training to train a CondInst~\cite{TianSC20} with the pseudo labels generated by BoxInst. However, the improvements are limited}
\label{fig:fine_mask_self_training}
\vspace{-12pt}
\end{figure}

Recently, a few works~\cite{BBTP,BBAM,BoxInst,Boxcaseg,BoxLevelSet,DiscoBox} explore weakly supervised instance segmentation with box annotations or low-level colors.
These weakly supervised methods can effectively train instance segmentation methods~\cite{HeGDG17,TianSC20,SOLOV2WangZKLS20} without pixel-wise or polygon-based annotations and obtain fine segmentation masks.
As shown in Fig.~\ref{fig:fine_mask_self_training}(a), BoxInst~\cite{BoxInst} can output a few high-quality segmentation masks and segment well on the object boundary, \eg, the person, even performs better than the ground-truth mask in details though other objects may be badly segmented.
Naturally, we wonder if the generated masks of box-supervised methods, especially the high-quality masks, could be qualified as pseudo segmentation labels to further improve the performance of weakly supervised instance segmentation.

To answer this question, we first employ the naive self-training to evaluate the performance of using box-supervised pseudo masks.
Given the generated instance masks from BoxInst, we propose a simple yet effective \textit{box-based pseudo mask assignment} to assign pseudo masks to ground-truth boxes.
And then we train the CondInst~\cite{TianSC20} with the pseudo masks, which has the same architecture with BoxInst and consists of a detector~\cite{FCOSTianSCH19} and a dynamic mask head.
Fig.~\ref{fig:fine_mask_self_training}(b) shows that using self-training brings minor improvements and fails to unleash the power of high-quality pseudo masks, which can be attributed to two obstacles, \ie, (1) the naive self-training fails to filter low-quality masks, and (2) the noisy pseudo masks hurt the training using fully-supervised pixel-wise loss. Besides, the multi-stage self-training is inefficient.

To address these problems, we present \name{}, an end-to-end training framework, which takes advantage of high-quality pseudo masks produced by box supervision.
\name{} is composed of a sophisticated \textit{Teacher} and a perturbed \textit{Student}, in which the teacher generates high-quality pseudo instance masks along with the \textit{mask-aware confidence scores} to estimate the quality of masks.
Then the proposed \textit{box-based pseudo mask assignment} will assign the pseudo masks to the ground-truth boxes.
The student is normally optimized with the ground-truth boxes and pseudo masks through box-based loss and \textit{noise-aware pseudo mask loss}, and then progressively updates the teacher via Exponential Moving Average (EMA).
In contrast to the naive multi-stage self-training, \name{} is more simple and efficient.
The proposed mask-aware confidence score effectively reduces the impact of low-quality masks.
More importantly, pseudo labeling can mutually improve the student and further enforce the teacher to generate higher-quality masks, hence pushing the limits of the box supervision.
\name{} can serve as a general training paradigm and is agnostic to the methods for instance segmentation.

To benchmark the proposed \name{}, we adopt CondInst~\cite{TianSC20} as the basic segmentation method. 
On the challenging COCO dataset~\cite{COCOLinMBHPRDZ14}, \name{} surprisingly achieves $35.0$ and $36.5$ mask AP based on ResNet-50~\cite{HeZRS16} and ResNet-101 respectively, which remarkably outperforms the counterparts.
We provide extensive experiments on PASCAL VOC and Cityscapes to demonstrate its effectiveness and generalization ability.
Furthermore, \name{} with Swin Transformer~\cite{SwinTransformer} obtains 40.6 mask AP as a weakly approach for instance segmentation.

Overall, the contribution can be summarized as follows:
\begin{itemize}
    \vspace{-5pt}
    \item We solve the box-supervised instance segmentation problem from a new perspective, \ie, self-training with pseudo masks, and illustrate its effectiveness.
    % \vspace{-5pt}
    \item We present \name{}, a simple yet effective framework, which leverages pseudo masks with the mask-aware confidence score and noise-aware pseudo masks loss. Besides, we propose a pseudo mask assignment to assign pseudo masks to ground-truth boxes.
    % \vspace{-5pt}
    \item We improve the weakly supervised instance segmentation by large margins and bridge the gap between box-supervised and mask-supervised methods, \eg, \name{} achieves $36.5$ mask AP on COCO compared to $39.1$ AP obtained by CondInst.
\end{itemize}

\section{Related Work}
\paragraph{Instance Segmentation.}
Methods for instance segmentation can be roughly divided into two groups, \ie, single-stage methods and two-stage methods.
Single-stage methods~\cite{YolactBolyaZXL19,TianSC20,PolarMaskXieSSWLLSL20,MEInstZhangTSYY20} tend to adopt single-stage object detectors~\cite{SSD,FCOSTianSCH19}, to localize and recognize objects, and then generate segmentation masks through object enmbeddings or dynamic convolution~\cite{DynamicConvChenDLCYL20}.
Wang \etal{} present box-free SOLO~\cite{SOLOWangKSJL20} and SOLOv2~\cite{SOLOV2WangZKLS20}, which are independent of object detectors.
SparseInst~\cite{sparseinst} and YOLACT~\cite{YolactBolyaZXL19}, aiming for real-time inference, achieve great trade-off between speed and accuracy.
Two-stage methods~\cite{HeGDG17,MSRCNNHuangHGHW19,PointRendKirillovWHG20,BMaskChengWH020} adopt bounding boxes from object detectors and RoIAlign~\cite{HeGDG17} to extract the RoI (region-of-interest) features for object segmentation, \eg, Mask R-CNN~\cite{HeGDG17}.
Several methods~\cite{MSRCNNHuangHGHW19,BMaskChengWH020,PointRendKirillovWHG20} based on Mask R-CNN are proposed to refine the segmentation masks for high-quality instance segmentation.
Recently, many approaches~\cite{DETRCarionMSUKZ20,queryinst,maskformer,mask2former,solq,knet} based on transformers~\cite{VaswaniSPUJGKP17,VIT} or the Hungarian algorithm~\cite{HungarianStewartAN16} have made great progress in instance segmentation.
\paragraph{Weakly Supervised Instance Segmentation.}
Considering the huge cost of labeling instance segmentation, weakly supervised instance segmentation using image-level labels or bounding boxes gets lots of attention.
Several methods~\cite{PRM,iam_zhu,AhnCK19,ArunJK20} exploit image-level labels to generate pseudo masks from activation maps.
Khoreva et.al.~\cite{KhorevaBH0S17} propose to generate pseudo masks with GrabCut~\cite{RotherKB04} from given bounding boxes.
% BBTP~\cite{BBTP} brings multiple instance learning (MIL) into Mask R-CNN for weakly supervised instance segmentation.
BoxCaseg~\cite{Boxcaseg} leverages a saliency model to generate pseudo object masks for training Mask R-CNN along with the multiple instance learning (MIL) loss.
Recently, many box-supervised methods~\cite{BBTP,BoxInst,DiscoBox,BoxLevelSet} combines the MIL loss or pairwise relation loss from low-level features obtain impressing results with box annotations.
In comparison with BoxInst~\cite{BoxInst}, \name{} inherits the box supervision~\cite{BoxInst} but concentrates more on the novel training paradigm and exploiting noisy pseudo masks for high-performance box-supervised instance segmentation with box annotations.
Different from DiscoBox~\cite{DiscoBox} based on mean teacher~\cite{mean_teacher}, \name{} aims at a simple yet effective training framework with obtaining high-quality pseudo masks and learning from noisy masks.
\paragraph{Semi-supervised Learning.} 
Pseudo labeling~\cite{selftraining_fra,UPS,BachmanAP14} and consistency regularization~\cite{UDA,FixMatch,SajjadiJT16,temporal_ensemble,MixMatch} have greatly advanced the semi-supervised learning, which enables the training on large-scale unlabeled datasets.
Recently, semi-supervised learning has been widely used in object detection~\cite{STAC,softteacher,unbiased_teacher} and semantic segmentation~\cite{semi_seg_cps,semi_simple_base,UUPL} and demonstrated its effectiveness.
Motivated by high-quality masks from box supervision, we adopt the successful pseudo labeling and consistency regularization to develop a new training framework for weakly supervised instance segmentation.
Compared to \cite{Ge2022PointTeachingWS} which has similar motivation but aims for semi-supervised object detection with labeled images and extra point annotations, \name{} addresses box-supervised instance segmentation with box-only annotations. 
Compared to \cite{Huang2022W2NSF,salvage} which adopt multi-stage training and combine weakly supervised and semi-supervised learning, \name{} is a one-stage framework without pre-trained labelers.

\section{Naive Self-Training with Pseudo Masks}
\paragraph{Revisiting Box-supervised Methods.}
\label{revisit}
Note that \textit{box-only} annotations is sufficient to train an object detector, which can accurately localize and recognize objects.
Box-supervised methods~\cite{BoxInst,BoxLevelSet,DiscoBox} based on object detectors mainly exploit two exquisite losses to supervise mask predictions, \ie, the multiple instance learning (MIL) loss and the pairwise relation loss.
Concretely, according to the bounding boxes, the MIL loss can determine the positive and negative bags of pixels of the predicted masks.
Pairwise relation loss concentrates on the local relations of pixels from low-level colors or features, in which neighboring pixels have the similar color will be regarded as a positive pair and should output similar probabilities.
The MIL loss and pairwise relation loss enables the box-supervised methods to produce the complete segmentation masks, and even some high-quality masks with fine details.

\paragraph{Naive Self-Training.}
\label{naive_self_training}
% \subsection{Naive Self-Training}
Considering that the box-supervised methods can produce some high-quality masks without mask annotations, we adopt self-training to utilize the high-quality masks as pseudo labels to train an instance segmentation method with full supervision.
Specifically, we adopt the successful BoxInst~\cite{BoxInst} to generate pseudo instance masks on the given dataset $\mathbb{X}=\{\mathcal{X}, \mathcal{B}^g\}$, which only contains the box annotations.
For each input image $\mathcal{X}$, let $\{\mathcal{B}^p, \mathcal{C}^p, \mathcal{M}^p\}$ denote the predicted bounding boxes, confidence scores, and predicted instance masks, respectively.
We propose a simple yet effective \textit{Box-based Pseudo Mask Assignment} algorithm in Alg.~\ref{alg:pseudo_labeling} to assign the predicted instance masks to the box annotations via the confidence scores and intersection-over-union (IoU) between ground-truth boxes $\mathcal{B}^g$ and predicted boxes $\mathcal{B}^p$.
The hyper-parameters $\tau_{\text{iou}}$ and $\tau_c$ are set to $0.5$ and $0.05$, respectively.
The assigned instance masks will be rectified by removing the parts beyond the bounding boxes.
Then, we adopt the dataset $\hat{\mathbb{X}}=\{\mathcal{X}, \mathcal{B}^g,\mathcal{M}^g \}$ with pseudo instance masks to train an approach, \eg, CondInst~\cite{TianSC20}.
\paragraph{Naive Self-Training is Limited.} 
Fig.~\ref{fig:fine_mask_self_training}(b) and Tab.~\ref{tab:ablation_self_training} provide the experimental results of using naive self-training pseudo masks.
Compared to the pseudo labeler, using self-training brings minor improvements and even fails to surpass the pseudo labeler.
We attribute the limited performance to two issues, \ie, the naive self-training fails to exclude low-quality masks and the fully-supervised loss is sensitive to the noisy pseudo masks.
% \RestyleAlgo{ruled}
\vspace{-8pt}
\begin{algorithm}
\LinesNumbered
% \algsetup{linenosize=\scriptsize} \small
\caption{Box-based Pseudo Mask Assignment}\label{alg:pseudo_labeling}
\SetKwInOut{Parameter}{Parameter}
\KwIn{ predicted boxes $\mathcal{B}^p \!\in\! \mathbb{R}^{N\!\times\!4}$, predicted masks $\mathcal{M}^p \!\in\! \mathbb{R}^{N\!\times\!H\!\times\!W}$, confidence score $\mathcal{C}^p \!\in\! \mathbb{R}^{N}$, ground-truth boxes $\mathcal{B}^g \!\in \!\mathbb{R}^{K\!\times\!4}$.}
\Parameter{IoU threshold $\tau_{\text{iou}}$, confidence threshold $\tau_c$.}
\KwOut{assigned pseudo masks $\mathcal{M}^g \!\in\! \mathbb{R}^{K\!\times\!H\!\times\!W}$.}
Initialize output masks $\mathcal{M}^g$ with empty ($0$), assignment index $A \!\in\! \mathbb{R}^{K}$ with $-1$\;
Filter the predictions by the confidence threshold $\tau_c$\;
Sort the confidence score $\mathcal{C}^p$ in descending order with output indices $S \in \mathbb{N}^{N}$\;
\ForEach{{\rm prediction} $i$ in $S$} {
  Initialize $u \leftarrow -1$, $v \leftarrow -1$\;
  \For{$j = 1$ \KwTo $K$}{
    $iou_{ij}=\texttt{ComputeIoU}(\mathcal{B}^p_i,\mathcal{B}^g_j)$\;
    \If{$A_j > 0$}{
        continue\;
    }
    \If{$\text{iou}_{ij} \geq \tau_{\text{iou}}$ \AND $\text{iou}_{ij} \geq u$} {
        $u \leftarrow \text{iou}_{ij}$, $v \leftarrow i$\;
    }
    \If{$v > 0$} {
        Assign mask $\mathcal{M}^p_i$ to mask $\mathcal{M}^g_v$ \;
        $A_j \leftarrow i$\;
    }
  }
}
\end{algorithm}
\vspace{-10pt}

\section{BoxTeacher}
In this section, we present \name{}, an end-to-end training framework, which aims to unleash the power of pseudo masks.
In contrast to multi-stage self-training, \name{}, consisting of a teacher and a student, simultaneously facilitates the training of the student and pseudo labeling of the teacher.
The mutual optimization is beneficial to both the teacher and the student, thus leading to higher performance for box-supervised instance segmentation.
% In this section, 

\begin{figure*}
    \centering
    \includegraphics[width=0.85\linewidth]{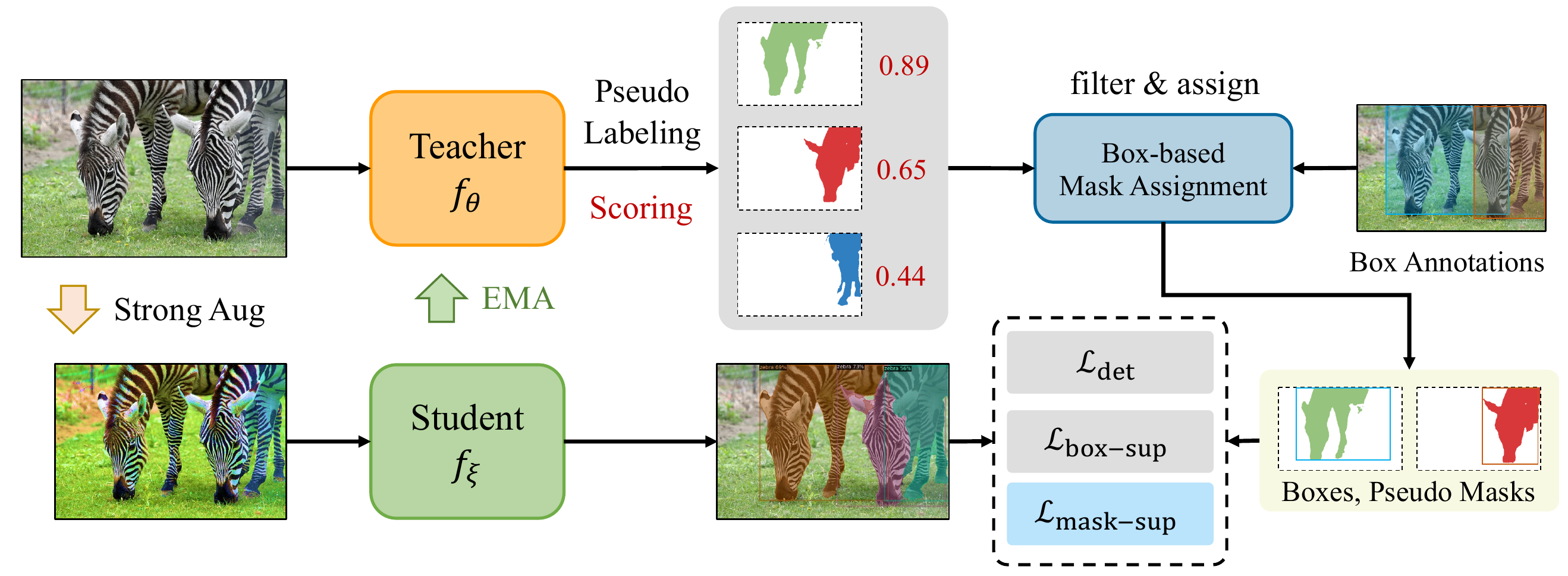}
    \vspace{-8pt}
    \caption{\textbf{The Architecture of \name{}.} Images are firstly fed into the \textit{Teacher} to obtain the pseudo masks and estimate the quality of masks. Then the box-based mask assignment filters and assigns pseudo masks to box annotations. The \textit{Student} adopt the augmented images (\ie, random scale or color jittering) and pseudo masks to update the parameters by gradient descent and then update the Teacher with exponential moving average (EMA).}
    \label{fig:main_arch}
    \vspace{-5pt}
\end{figure*}

\subsection{Architecture}

The overall architecture of \name{} is depicted in Fig.~\ref{fig:main_arch}.
\name{} is composed of a teacher and 
a student, which shares the same model.
Given the input image, the teacher $f_{\theta}$ straightforwardly generates the predictions, including the bounding boxes, segmentation masks, and \textit{mask-aware confidence scores}. 
Similarly, we apply the \textit{box-based pseudo mask assignment} in Alg.~\ref{alg:pseudo_labeling} to assign the predicted masks to the ground-truth annotations.
Inspired by consistency regularization~\cite{UDA,FixMatch,STAC,unbiased_teacher}, we adopt strong augmentation for images (\eg, color jittering) fed into the student $f_{\xi}$ and the student is optimized under the box supervision and the mask supervision.
To acquire high-quality pseudo masks, we adopt the exponential moving average to gradually update the teacher from student~\cite{mean_teacher}, \ie, $f_{\theta} \leftarrow \alpha\cdot f_{\theta} + (1-\alpha)\cdot f_{\xi}$ ($\alpha$ is empirically set $0.999$). 

\paragraph{Mask-aware Confidence Score.} 
Considering that the generated pseudo masks are noisy and unreliable, which may hurt the performance, we define the mask-aware confidence score to estimate the quality of the pseudo masks.
Inspired by~\cite{SOLOWangKSJL20}, we denote $m^b_i \in \mathbb{R}^{H\times W}$ and $m_i \in \mathbb{R}^{H\times W}$ as the box-based binary masks and sigmoid probabilities of the $i$-th pseudo mask with the detection confidence $c_i$, the mask-aware confidence score $s_i$ is defined as follows:
\begin{equation}
\label{eq:maskscore_definition}
    \small
    s_i = \sqrt{c_i \cdot \frac{\sum^{H,W}_{x,y}\mathbbm{1}(m_{i,x,y}>\tau_m)\cdot m_{i,x,y}\cdot m^b_{i,x,y}}{\sum^{H,W}_{x,y}\mathbbm{1}(m_{i,x,y}>\tau_m)\cdot m^b_{i,x,y}}},
\end{equation}
where $\mathbbm{1}(\cdot)$ is the indicator function, $\tau_m$ is the threshold for binary masks and set to $0.5$.
The mask-aware score calculates the average probability score of the positive masks inside the ground-truth boxes, and the higher average probability means more confident pixels in the mask.
In addition, we explore several kinds of quality scores and compare with the mask-aware score in experiments.

\paragraph{Training Loss.}

\name{} can be end-to-end optimized with box annotations and the generated pseudo masks.
The overall loss is defined as: $\mathcal{L}\!=\!\mathcal{L}_{\text{det}}\!+\!\mathcal{L}_{\text{box-sup}}\!+\!\mathcal{L}_{\text{mask-sup}}$, which consists of the standard detection loss $\mathcal{L}_{\text{det}}$, the box-supervised loss $\mathcal{L}_{\text{box-sup}}$, and the mask-supervised loss $\mathcal{L}_{\text{mask-sup}}$.
We inherit the detection loss defined in FCOS~\cite{TianSC20}, and we follow previous works~\cite{BoxInst,BBTP,DiscoBox} to adopt the max-projection loss and the color-based pairwise relation loss~\cite{BoxInst} for box-supervised mask loss $\mathcal{L}_{\text{box-sup}}$.

\subsection{Noise-aware Pseudo Mask Loss}
The goal of \name{} is to take advantage of high-quality pseudo masks in a fully supervised manner while reduce the impact of the noisy or low-quality instance masks.
To this end, we present the noise-aware pseudo mask loss in Eq.~\eqref{pseudo_loss}.
Ideally, \name{} can leverage the pseudo masks to calculate the fully-supervised pixel-wise segmentation loss, \eg, dice loss~\cite{dice}.
Besides, we also propose a novel \textit{noise-reduced mask affinity loss} $\mathcal{L}_{\text{affinity}}$ to enhance the pixel-wise segmentation with neighboring pixels.
Further, we employ the proposed mask-aware confidence scores $\{s_i\}$ as weights for the pseudo mask loss, which adaptively scales the weights for pseudo masks of different qualities.
The total pseudo mask loss is defined as follows:
\begin{equation}
    \label{pseudo_loss}
    \small
   \mathcal{L}_{\text{mask-sup}}\!=\! \frac{1}{N_p}\!\sum_{i=1}^{N_p}s_i\cdot(\lambda_{p}\mathcal{L}_{\text{pixel}}(m^p_i,m^g_i) + \lambda_{a}\mathcal{L}_{\text{affinity}}(m^p_i,m^g_i)),
   % + \lambda_3\mathcal{L}_{\text{avg}}(m^p_i,m^g_i)), 
\end{equation}
where $m_i^p$ and $m_i^g$ denotes the $i$-th predicted masks and pseudo masks, $N_p$ denotes the number of valid pseudo masks, $\lambda_{\text{pixel}}$ and $\lambda_{\text{affinity}}$ are set to $0.5$ and $0.1$ respectively.
To stabilize the training, we adopt a linear warmup strategy for pseudo mask loss at the beginning of the training, \ie, the first $10$k iterations.

\paragraph{Noise-reduced Mask Affinity Loss.} 
considering pseudo masks contain much noise while neighboring pixels in local regions (\eg, $3\!\times\!3$) tend to have similar semantics or labels, we exploit label affinities among pixels within local regions to alleviate label noise.
Given the $i$-th pixel sigmoid probability $g_i$ of the pseudo segmentation, we first calculate the refined pixel probability $\tilde{g_i}$ with its neighboring pixels, which is defined as follows:
\begin{equation}
    \tilde{g_i} = \frac{1}{2}(g_i + \frac{1}{|\mathcal{P}|}\sum_{j \in \mathcal{P}} g_j),
\end{equation}
where $\mathcal{P}$ denotes the set of neighboring pixels, \eg, a $3\times3$ region.
This refinement can reduce the outliers and enhance the pixels with local context.
Then, We present a simple noise-reduced mask affinity loss and define the affinity $\mu_{ij}$ between $i$-th and $j$-th pixels as follows:
\begin{equation}
\small
\mu_{ij} = \tilde{g_i}\cdot\tilde{g_j} + (1 - \tilde{g_i}) \cdot (1-\tilde{g_j}),
\end{equation}
where $\tilde{g_i}$ and $\tilde{g_j}$ are refined pixels which encode the local context. 
Then the \textit{noise-reduced mask affinity loss} for $i$-th pixel is defined as follows:
\begin{equation}
    \small
    \mathcal{L}_{\text{affinity}}\!=\!-\frac{\sum\limits_{j\in\mathcal{P}}\!\mathbbm{1}(\mu_{ij}\!>\!\tau_a)(\log(p_i\cdot p_j)\!+\!\log((1\!-\!p_i)\!\cdot\!(1\!-\!p_j)))}{\sum\limits_{j\in\mathcal{P}}\mathbbm{1}(\mu_{ij}>\tau_a)},
\end{equation}
where $j\in \mathcal{P}$ are the neighboring pixels of the $i$-th pixel and $\tau_a$ is set to 0.5 as default. $p_i$ denotes the $i$-th pixel of the predicted mask.

\section{Experiments}

In this section, we mainly evaluate the proposed \name{} on the COCO dataset~\cite{COCOLinMBHPRDZ14}, the Cityscapes dataset~\cite{cordts2016cityscapes}, and the PASCAL VOC 2012 dataset~\cite{pascalvoc}, and provide extensive ablations to analyze the proposed method.
% We also refer the readers to the Appendix for additional ablations and visualizations.
\paragraph{Datasets.} The COCO dataset contains $80$ categories and $110k$ images for training, $5k$ images for validation, and $20k$ images for testing.
The Cityscapes dataset, aiming for perception in driving scenes, consists of $5000$ street-view high-resolution images, in which $2975$, $500$, and $1525$ images are used for training, validation, and testing, respectively.
Foreground objects in Cityscapes are categorized into $8$ classes and fine-annotated with pixel-wise segmentation labels instead of polygons adopted in COCO, thus making the labeling process much costly.
The PASCAL VOC 2012 dataset has $20$ categories and and consists of $10582$ images for training amd 1449 images for validation.
For weakly supervised instance segmentation, we only keep the bounding boxes and ignore the segmentation masks during training.
\paragraph{Implementation Details.}
The proposed \name{} is implemented based on PyTorch~\cite{pytorch} and
we mainly adopt CondInst~\cite{TianSC20} as the meta method for instance segmentation.
The backbone networks are initialized with the ImageNet-pretrained weights and the BatchNorm layers are frozen.
All \name{} models are trained over $8$ GPUs.

\paragraph{Data Augmentation.}
For images input to the student, we adopt random horizontal flip and the multi-scale augmentation which randomly resizes images from $640$ to $800$ as the basic augmentation.
In addition, we randomly apply color jittering, grayscale, and Gaussian blur for stronger augmentation.
While the images fed into the teacher are fixed to $800\times1333$ without perturbation.

\subsection{Instance Segmentation on COCO}
\paragraph{Experimental Setup.}
\label{expr_details}
Following the training recipes~\cite{FCOSTianSCH19,TianSC20,BoxInst}, \name{} is trained with $16$ images per batch.
Unless specified, we adopt the standard $1\times$ schedule ($90k$ iterations) ~\cite{HeGDG17,wu2019detectron2} with the SGD and the initial learning rate $0.01$.
For comparisons with the state-of-art methods, we scale up the learning schedule to $3\times$ ($270k$ iterations).
% For images input to the student, we adopt the multi-scale augmentation which randomly resizes images from $640$ to $800$.
\paragraph{Main Results.}
Tab.~\ref{tab:main_experiments} shows the main results on COCO \texttt{test-dev}.
In comparison with other state-of-the art methods, we evaluate the proposed \name{} with different backbone networks, \ie, ResNet~\cite{HeZRS16} and Swin Transformer~\cite{SwinTransformer}, and under different training schedules, \ie, $1\times$ and $3\times$.
It's clear that \name{} with ResNet-50 achieves $32.9$ mask AP, which  outperforms other box-supervised methods~\cite{DiscoBox,BoxInst} even with longer schedules.
Compared to recent BoxInst~\cite{BoxInst}, BoxLevelSet~\cite{BoxLevelSet} and DiscoBox~\cite{DiscoBox}, \name{} significantly brings about $3.0$ mask AP improvements based on ResNet-50 and ResNet-101 under the same setting.
Remarkably, \name{} also bridges the gap between {mask-supervised} methods and {box-supervised} methods, \eg, the gap between \name{} based on ResNet-101 and CondInst is reduced to $2.6$ AP.
With the stronger backbones, \eg, Swin Transformer~\cite{SwinTransformer}, \name{} can surprisingly obtain $40.6$ mask AP on COCO dataset, which is highly competitive as a weakly supervised method for instance segmentation. 

\begin{table*}
    \caption{\textbf{Results on COCO Instance Segmentation.} Comparisons with state-of-the-art methods on COCO \texttt{test-dev}. With the same backbone or learning schedule, \name{} surprisingly surpasses the counterparts by large margins (more than $2.0$ mask AP). $\dagger$: trained without strong augmentation. `R-50' and `R-101' denote ResNet-50 and ResNet-101, and `DCN' denotes deformable convolution~\cite{dcn,dcnv2}.}
    \vspace{-5pt}
    \centering
    \renewcommand{\tabcolsep}{6pt}
    \renewcommand\arraystretch{1.1}
    % \scalebox{0.9}{
    \small
    \begin{tabular}{l|l|c|ccc|ccc}
    % \toprule
    Method & Backbone & Schedule & \apm & \apl\\
    \tline
    \multicolumn{8}{l}{\textit{Mask-supervised methods.}} \\
    \hline
    Mask R-CNN~\cite{HeGDG17} & R-50-FPN & $1\times$ & $35.5$ & $57.0$ & $37.8$ & $19.5$ & $37.6$ & $46.0$ \\
    CondInst~\cite{TianSC20} & R-50-FPN & $1\times$ & $35.9$ & $57.0$ & $38.2$ & $19.0$ & $38.6$ & $46.7$ \\
    CondInst~\cite{TianSC20} & R-50-FPN & $3\times$ & $37.7$ & $58.9$ & $40.3$ & $20.4$ & $40.2$ & $48.9$ \\
    CondInst~\cite{TianSC20} & R-101-FPN & $3\times$ & $39.1$ & $60.9$ & $42.0$ & $21.5$ & $41.7$ & $50.9$ \\
    SOLO~\cite{SOLOWangKSJL20} & R-101-FPN & $6\times$ & $37.8$ & $59.5$ & $40.4$ & $16.4$ & $40.6$ & $54.2$ \\
    SOLOv2~\cite{SOLOWangKSJL20}  & R-101-FPN & $6\times$ & $39.7$ & $60.7$ & $42.9$ & $17.3$ & $42.9$ & $57.4$ \\
    \tline
    \multicolumn{8}{l}{\textit{Box-supervised methods.}} \\
    \hline
    BoxInst~\cite{BoxInst} & R-50-FPN & $3\times$ & $32.1$ & $55.1$ & $32.4$ & $15.6$ & $34.3$ & $43.5$\\
    DiscoBox~\cite{DiscoBox} & R-50-FPN & $3\times$ & $32.0$ & $53.6$ & $32.6$ & $11.7$ & $33.7$ & $48.4$ \\
    \hline
    \name{}$^\dagger$ & R-50-FPN & $1\times$ & $32.9$ & $54.1$ & $34.2$ & $17.4$ & $36.3$ & $43.7$ \\
    \name{} & R-50-FPN & $3\times$ & $35.0$ & $56.8$ & $36.7$ & $19.0$ & $38.5$ & $45.9$\\
    \hline
    \hline
    BBTP~\cite{BBTP} & R-101-FPN & $1\times$ & $21.1$ & $45.5$ & $17.2$ & $11.2$ & $22.0$ & $29.8$ \\
    BBAM~\cite{BBAM} & R-101-FPN & $1\times$ & $25.7$ & $50.0$ & $23.3$ & - & - & - \\
    BoxCaseg~\cite{Boxcaseg} & R-101-FPN & $1\times$ & $30.9$ & $54.3$ & $30.8$ & $12.1$ & $32.8$ & $46.3$ \\
    BoxInst~\cite{BoxInst} & R-101-FPN & $3\times$ & $33.2$ & $56.5$ & $33.6$ & $16.2$ & $35.3$ & $45.1$\\
    BoxLevelSet~\cite{BoxLevelSet} & R-101-FPN & $3\times$ & $33.4$ & $56.8$ & $34.1$ & $15.2$ & $36.8$ & $46.8$ \\
    BoxLevelSet~\cite{BoxLevelSet} & R-101-DCN-FPN & $3\times$ & $35.4$ & $59.1$ & $36.7$ & $16.8$ & $38.5$ & $51.3$ \\
    DiscoBox~\cite{DiscoBox} & R-101-DCN-FPN & $3\times$ & $35.8$ & $59.8$ & $36.4$ & $16.9$ & $38.7$ & $52.1$ \\
    \hline
    \name{} & R-101-FPN & $3\times$ & $36.5$ & $59.1$ & $38.4$ & $20.1$ & $40.2$ & $47.9$ \\
    \name{} & R-101-DCN-FPN & $3\times$ & $37.6$ & $60.3$ & $39.7$ & $21.0$ & $41.8$ & $49.3$\\
    \name{} & Swin-Base-FPN & $3\times$ & $40.6$ & $65.0$ & $42.5$ & $23.4$ & $44.9$ & $54.2$ \\
    \end{tabular}
    \vspace{-8pt}
    \label{tab:main_experiments}
\end{table*}

\subsection{Instance Segmentation on PASCAL VOC}
\paragraph{Experimental Setup.}
\name{} is trained for $15k$ iterations with 16 images per batch.
Following previous works~\cite{BoxLevelSet,DiscoBox,BBAM}, we report COCO-style AP and AP under 4 thresholds, \ie, $\{0.25, 0.5, 0.70, 0.75\}$.
\paragraph{Main Results.}
Tab.~\ref{tab:main_voc_experiments} shows the comparisons with the state-of-the-art methods on PASCAL VOC 2012.
Compared to recent box-supervised methods~\cite{BoxInst,BoxLevelSet,DiscoBox}, the proposed \name{} achieves better results under different IoU thresholds, which remarkably outperforms BoxInst and DiscoBox by large margins.
Notably, \name{} obtains significant improvements under higher IoU thresholds.

\begin{table}
    \caption{\textbf{Results on PASCAL VOC.} Comparisons with the state-of-the-art methods on PASCAL VOC 2012 \texttt{val}. All methods adopt \textit{box-only} annotations.}
    \vspace{-3pt}
    \centering
    \renewcommand{\tabcolsep}{3pt}
    \renewcommand\arraystretch{1.1}
    % \scalebox{1.0}{
    \small
    \begin{tabular}{l|l|ccccc}
    Method & Backbone & AP & AP$_{25}$ & AP$_{50}$ & AP$_{70}$ & AP$_{75}$ \\
    \tline
    SDI~\cite{KhorevaBH0S17} & VGG-16 & - & -  & $44.8$ & - & $16.3$ \\
    BoxInst~\cite{BoxInst} & R-50 & $34.3$ & - & $59.1$ & - & $34.2$ \\
    DiscoBox~\cite{DiscoBox} & R-50 & - & $71.4$ & $59.8$ & $41.7$ & $35.5$ \\
    BoxLevelSet~\cite{BoxLevelSet} & R-50 & $36.3$ & $76.3$ & $64.2$ & $43.9$ & $35.9$\\
    \hline
    \name{} & R-50 & $38.6$ & $77.6$ & $66.4$ & $46.1$ & $38.7$ \\
    \hline
    \hline
    BBTP~\cite{BBTP} & R-101 & - & $75.0$ & $58.9$ & $30.4$ & $21.6$ \\ 
    Arun \etal~\cite{ArunJK20} & R-101 & - & $73.1$ & $57.7$ & $33.5$ & $31.2$ \\   
    BBAM~\cite{BBAM} & R-101 & - & $76.8$ & $63.7$ & $39.5$ & $31.8$ \\
    BoxInst~\cite{BoxInst} & R-101 & $36.4$ & - & $61.4$ & - & $37.0$ \\
    DiscoBox~\cite{DiscoBox} & R-101 & - & $72.8$ & $62.2$ & $45.5$ & $37.5$ \\
    BoxLevelSet~\cite{BoxLevelSet} & R-101 & $38.3$ & $77.9$ & $66.3$ & $46.4$ & $38.7$\\
    \hline
    \name{} & R-101 & $40.3$ & $78.4$ & $67.8$ & $48.0$ & $41.3$ \\
    \end{tabular}
    % \vspace{2pt}
    \label{tab:main_voc_experiments}
    \vspace{-15pt}
\end{table}

\subsection{Instance Segmentation on Cityscapes}
\paragraph{Experimental Setup.}
Following previous methods~\cite{HeGDG17,TianSC20}, we train all models for $24k$ iterations with $8$ images per batch.
The initial learning rate is $0.005$.
Cityscapes contains high-resolution images ($2048\!\times\!1024$), and we randomly resize images from $800$ to $1024$ for the student and keep the original size for the teacher during training.
In addition, we also adopt the COCO pre-trained models ($1\times$ schedule) to initialize the weights for higher performance.
\paragraph{Main Results.}
Tab.~\ref{tab:main_city_experiments} shows the evaluation results on Cityscapes \texttt{val}.
\name{} outperforms previous box-supervised methods significantly, especially with the COCO pre-trained weights.
Though performance gap between fully supervised methods and weakly supervised methods become larger than that in COCO, the human labour of labeling pixel-wise segmentation for a high-resolution Cityscapes image is much costly (90 minutes per image).
And we hope future research can bridge the gap between box-supervised methods and mask-supervised methods for high-resolution images. 
\begin{table}
    \caption{\textbf{Results on Cityscapes Instance Segmentation.} Comparisons with state-of-the-art methods for mask AP on Cityscapes \texttt{val}. $^\dagger$: our re-produced results on Cityscapes based on the public code.`\texttt{fine}' denotes the Cityscapes \texttt{train} with fine annotations while `\texttt{fine}+COCO' denotes using COCO pre-trained weights. For box-supervised methods, we remove the fine-grained mask annotations in Cityscapes.}
    \vspace{-8pt}
    \centering
    \renewcommand\arraystretch{1.1}
    % \scalebox{1.0}{
    \small
    \begin{tabular}{l|l|cc}
    % \toprule
    Method  & Data & AP & AP$_{50}$ \\
    \tline
    \multicolumn{4}{l}{\textit{Mask-supervised methods.}} \\
    \hline
    Mask R-CNN~\cite{HeGDG17} & \texttt{fine} & $31.5$ & - \\
    CondInst~\cite{TianSC20}  & \texttt{fine} & $33.0$ & $59.3$ \\
    CondInst~\cite{TianSC20} & \texttt{fine} + COCO & $37.8$ & $63.4$ \\
    \hline
    \multicolumn{4}{l}{\textit{Box-supervised methods.}} \\
    \hline    
    BoxInst$^\dagger$~\cite{BoxInst}  & \texttt{fine} & $19.0$ & $41.8$ \\ 
    BoxInst$^\dagger$~\cite{BoxInst}  & \texttt{fine} + COCO & $24.0$ & $51.0$ \\
    BoxLevelSet$^\dagger$~\cite{BoxLevelSet}  & \texttt{fine} & $20.7$ & $43.3$ \\
    BoxLevelSet$^\dagger$~\cite{BoxLevelSet}  & \texttt{fine} + COCO & $22.7$ & $46.6$\\
    \hline
    \name{} & \texttt{fine} & $21.7$ & $47.5$ \\
    \name{} & \texttt{fine} + COCO & $26.8$ & $54.2$\\
    % BoxTeacher & R-50-FPN & \texttt{fine} + COCO & $29.2$ & $58.1$\\
    \end{tabular}
    % \vspace{2pt}
    \label{tab:main_city_experiments}
    \vspace{-15pt}
\end{table}

\subsection{Ablation Experiments}

\begin{table*}
\centering
\begin{minipage}[t]{0.30\textwidth}
    \caption{\textbf{Pseudo Mask Loss.} We evaluate the effects of different loss for \name{}.}
    \renewcommand{\tabcolsep}{4pt}
    \renewcommand\arraystretch{1.1}
    \small
    \scalebox{1.0}{
    \begin{tabular}{lc|ccc}
    $\mathcal{L}_{\text{pixel}}$ & $\mathcal{L}_{\text{affinity}}$ & \aps  \\
    \tline
    \gt{\xmark} & \gt{-} & \gt{$30.7$} & \gt{$52.5$} & \gt{$31.2$} \\
    bce & - & $28.9$ & $49.2$ & $29.5$ \\
    dice & -  & $31.8$ & $53.1$ & $32.8$ \\
    \hline
    % \xmark  & - & \checkmark  & $31.4$ & $52.9$ & $32.2$ \\
    % dice & - & \checkmark  & $31.9$ & $53.1$ & $33.0$\\
    dice & \checkmark  & $32.2$ & $53.5$ & $33.2$\\
    \end{tabular}}
    %\vspace{2pt}
    \label{tab:ablation_loss}
\end{minipage}
\hfill
%\end{table}
\begin{minipage}[t]{0.32\textwidth}
%\begin{table}
    \centering
    \small
    \caption{\textbf{Effect of the Weights of Pseudo Mask Loss.} We adopt $\lambda_{\text{pixel}}\!=\!0.5$ and $\lambda_{\text{affinity}}\!=\!0.1$ as the default setting.}
    \renewcommand{\tabcolsep}{6pt}
    \renewcommand\arraystretch{1.1}
    \scalebox{1.0}{
    \begin{tabular}{cc|ccc}
    $\lambda_{\text{pixel}}$ & $\lambda_{\text{affinity}}$ & \aps  \\
    \tline
    $0.1$ & - & $31.4$ & $53.0$ & $32.4$ \\
    \gt{$0.5$} & \gt{-} & \gt{$31.8$} & \gt{$53.1$} & \gt{$32.8$} \\
    $1.0$ & - & $31.5$ & $52.8$ & $32.3$ \\
    \hline
    $0.5$ & $0.1$ & $32.2$ & $53.5$ & $33.2$ \\
    $0.5$ & $0.5$ & $31.7$ & $52.8$ & $32.8$\\
    \end{tabular}}
    % \vspace{2pt}
    \label{tab:ablation_loss_weight}
\end{minipage}
\hfill
\begin{minipage}[t]{0.35\textwidth}
\centering
    \captionof{table}{\textbf{Effects of Mask Score.} We evaluate different mask scores, and it shows that the mask-aware confidence performs better.}
    \centering
    \small
    \renewcommand{\tabcolsep}{6pt}
    \renewcommand\arraystretch{1.1}
    \scalebox{1.0}{
    \begin{tabular}{l|cccc}
    % \toprule
    Mask Score & \aps\\
    \tline
    \gt{\xmark}  & \gt{$32.2$} & \gt{$53.5$} & \gt{$33.2$} \\
    cls & $32.0$ & $53.5$ & $33.1$\\
    iou & $32.2$ & $53.5$ & $33.4$\\
    mean-entropy & $31.8$ & $53.3$ & $32.6$\\
    mask-aware & $32.6$ & $53.5$ & $33.8$\\
    \end{tabular}}
    \label{tab:ablation_mask_score}
\end{minipage}
\vspace{-5pt}
\end{table*}
\begin{table*}
    \caption{\textbf{Comparison with Naive Self-Training.} As discussed in Sec.~\ref{naive_self_training}, we leverage the pre-trained BoxInst to generate pseudo mask labels and assign the pseudo masks to the ground-truth boxes.
    Then we adopt the pseudo masks and train the CondInst with different schedules and backbones. $^\dagger$: the mask AP achieved by the pseudo labeler, \ie, BoxInst,  with \textit{box-only} annotations.
    $^\ddagger$: the ideal mask AP could be achieved by CondInst if trained with box annotations following BoxInst.}
    \vspace{-3pt}
    \centering
    \small
    \renewcommand{\tabcolsep}{6pt}
    \renewcommand\arraystretch{1.1}
    \scalebox{1.0}{
    \begin{tabular}{l|l|c|lc|c|ccc}
    % \toprule
    Method & Backbone & Schedule & Pseudo Label & AP$^\dagger$ & AP$^\ddagger$ & \aps\\
    \tline
    CondInst & R-50 & $1\times$ & BoxInst, R-50 & $30.7$ & $30.7$ & $31.0$ & $53.1$ & $31.6$ \\
    CondInst & R-50 & $3\times$ & BoxInst, R-50 & $30.7$ & $31.8$ & $31.3$ & $53.8$ & $31.7$ \\
    CondInst & R-50 & $3\times$ & BoxInst, R-101 & $33.0$ & $31.8$ & $32.5$ & $54.9$ & $33.2$\\
    CondInst & R-101 & $3\times$ & BoxInst, R-101 & $33.0$ & $33.0$ & $32.9$ & $55.4$ & $33.7$ \\
    \hline
    \hline
    \name{} & R-50 & $1\times$ & End-to-End & - & - & $32.6$ & $53.5$ & $33.8$ \\
    \name{} & R-50 & $3\times$ & End-to-End & - & - & $34.2$ & $56.0$ & $35.4$ \\
    \name{} & R-101 & $3\times$ & End-to-End & - & -  & $35.2$ & $57.1$ & $36.8$ \\
    \end{tabular}}
    \vspace{-10pt}
    \label{tab:ablation_self_training}
\end{table*}

\begin{table}
    \caption{\textbf{The Effects of Data Augmentation.} We explore whether strong data augmentation will be beneficial to \name{}, which has been widely exploited in semi-supervised methods. We apply weak and strong augmentation to both CondInst and the proposed \name{}. AP$^b$ and AP$^m$ denote the AP for box and mask.}
    \centering
    \small
    \renewcommand\arraystretch{1.1}
    \begin{tabular}{lc|cc|cl}
    % \toprule
    Method & Schd. & weak & strong & AP$^{b}$ & AP$^{m}$\\
    \tline
    CondInst & $1\times$ & & & $39.6$ & $36.2$ \\
    CondInst & $1\times$ & \checkmark & & $39.6$ & $35.6^{-0.6}$ \\
    CondInst & $1\times$ & & \checkmark & $39.2$ & $35.3^{-0.9}$ \\
    \hline
    \name{} & $1\times$ & & & 39.4 & $32.6$ \\
    \name{} & $1\times$ & \checkmark & & $39.1$ & $32.4^{-0.2}$ \\
    \name{} & $1\times$ & & \checkmark & $38.8$ & $32.2^{-0.4}$ \\
    \hline
    \hline
    CondInst & $3\times$ & & & $41.9$ & $37.5$ \\
    CondInst & $3\times$ & & \checkmark & $42.0$ & $37.6^{+0.1}$ \\
    \hline
    \name{} & $3\times$ & & & $41.7$ & $34.2$ \\
    % & $34.2$ & $56.0$ & $35.4$
    \name{} & $3\times$ & & \checkmark & $41.8$ & $34.8^{+0.6}$ \\
    \end{tabular}
    \vspace{-10pt}
    \label{tab:ablation_data_aug}
\end{table}
\begin{table}
    \caption{\textbf{Ablations on Exponential Moving Average.} We evaluate the performance of BoxInst \textit{w/} or \textit{w/o} EMA to make it clear whether the improvements are brought by EMA in \name{}.}
    \centering
    \small
    \renewcommand\arraystretch{1.1}
    \scalebox{1.0}{
    \begin{tabular}{l|c|cccc}
    % \toprule
    Method & w/ EMA & AP$^{bbox}$ & \aps\\
    \tline
    BoxInst & \xmark & $39.3$ & $30.6$ & $52.2$ & $31.0$ \\
    BoxInst & \checkmark & $39.4$ & $30.7$ & $52.5$ & $31.2$ \\
    \end{tabular}}
    \vspace{-10pt}
    \label{tab:ablation_boxinst_ema}
\end{table}

\begin{figure*}
    \centering
    \includegraphics[width=\linewidth]{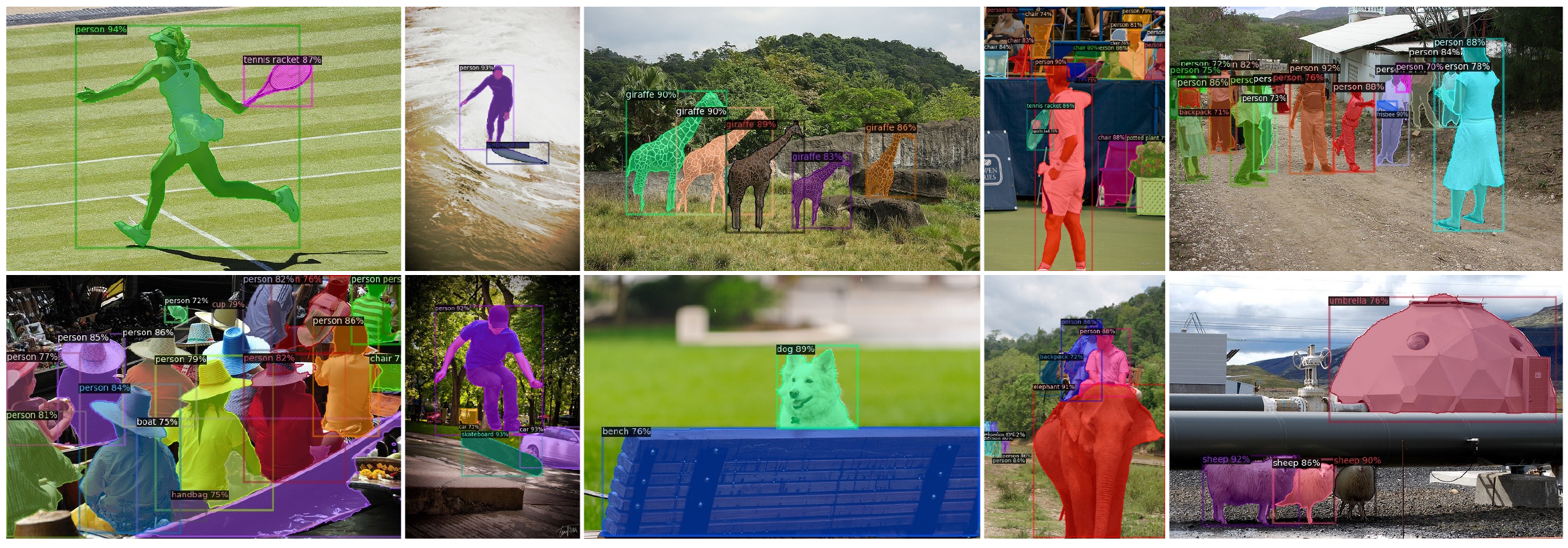}
    \vspace{-15pt}
    \caption{\textbf{Visualization Results.} We provide the visualization results of \name{} with ResNet-101 on the COCO \texttt{test-dev}. 
    The proposed \name{} can produce the high-quality segmentation results, even in some complicated scenes.}
    \label{fig:comparison_test}
    \vspace{-10pt}
\end{figure*}

\paragraph{Effects of Pseudo Mask Loss.}
In Tab.~\ref{tab:ablation_loss}, we explore the different pseudo mask loss for \name{}.
Firstly, we apply the box-supervised loss proposed in ~\cite{BoxInst} achieves $30.7$ mask AP (gray row).
As shown in Tab.~\ref{tab:ablation_loss},  directly applying binary cross entropy (bce) loss with pseudo masks leads to severe performance degradation, which can be attributed to the foreground/background imbalance and noise in pseudo masks.
Using dice loss to supervise the training with pseudo masks can bring significant improvements in comparison to the baseline.
% In addition, we adopt the weakly average projection loss proposed in ~\cite{freesolo}, which aims for coarse pseudo instance masks.
% Tab.~\ref{tab:ablation_loss} shows that average projection loss $\mathcal{L}_{\text{avg}}$ is inferior to the fully-supervised dice loss.
Adding mask affinity loss $\mathcal{L}_{\text{affinity}}$ provides $0.4$ AP gain based on the dice loss.
Moreover, we ablate the loss weights in pseudo mask loss in Tab.~\ref{tab:ablation_loss_weight}.
\paragraph{Effects of Mask-aware Confidence Score.}
Tab.~\ref{tab:ablation_mask_score} explores several different scores to estimate the quality of pseudo masks in an unsupervised manner, \ie, (1) classification scores (cls), (2) matched IoU between predicted boxes and ground-truth boxes (iou), (3) mean entropy of the pixel probabilities of pseudo masks (mean-entropy: $s\!=\!1\!+\!\frac{1}{HW}\sum^{H,W}_{i,j}(p_{ij}\log{p_{ij}}\!+\!(1\!-\!p_{ij})\log(1\!-\!p_{i,j}))$), (4) the proposed mask-aware score (mask-aware).
As Tab.~\ref{tab:ablation_mask_score} shows, using the proposed mask-aware confidence score leads to better performance for \name{}.
Notably, measuring the quality of predicted masks is critical but challenging for leverage pseudo masks.
Accurate quality estimation can effectively reduce the impact of noisy masks

\paragraph{Comparisons with Self-Training Paradigm.}
We adopt the box-supervised approach, \ie, BoxInst~\cite{BoxInst}, to generate pseudo masks, which is pre-trained on COCO with \textit{box-only} annotations.
And then we assign the pseudo masks to the ground-truth boxes through the assignment Alg.~\ref{alg:pseudo_labeling}. 
As shown in Tab.~\ref{tab:ablation_self_training}, the improvements provided by self-training are much limited and the naive self-training even performs worse than the training with \textit{box-only} annotations, \eg, CondInst with R-50 and $3\times$ schedule obtains $31.3$ AP with pseudo masks, but inferior to the box-supervised version ($31.8$ AP).
Though the self-training scheme enables the supervised training with pseudo masks and achieves comparable performance, we believe the high-quality pseudo masks are not well exploited.
Significantly, \name{} achieves higher mask AP compared to both self-training, in an end-to-end manner without complicated steps or procedures for label generation.

\paragraph{Effects of the Strong Data Augmentation.}
In this study, we explore the effect of strong augmentation on the proposed \name{}, and apply augmentation to the input images of the student.
Specifically, we defined two levels of data augmentation in A.1 (in Appendix), \ie, strong augmentation and weak augmentation.
As Tab.~\ref{tab:ablation_data_aug} shows, both strong and weak augmentation hurt the performance of CondInst and \name{} under the $1\times$ training schedule.
Differently, \name{} is more robust to the augmentations as the mask AP drops $0.4$ compared to CondInst.
However, \name{} remarkably benefits more from the strong data augmentation when increasing the schedule to $3\times$.
In comparison to CondInst, \name{} with strong augmentation will enforce the consistency between the student and teacher.
Interestingly, Tab.~\ref{tab:ablation_data_aug} indicates that using strong augmentation is merely beneficial to the weakly supervised instance segmentation ($+0.6$ AP), but has no effect to the fully supervised object detection ($+0.1$AP), suggesting that consistency regularization might facilitate the learning from noisy pseudo masks.

\paragraph{Effects of Exponential Moving Average.}
To see whether EMA could partially bring some performance improvements, we re-train BoxInst with EMA to obtain the averaged model to evaluate the performance.
Tab.~\ref{tab:ablation_boxinst_ema} shows that applying EMA has little impact to the final performance, proving that the improvements of \name{} are mainly brought by the effects of pseudo masks.
% Besides, the L2 norm between 
\paragraph{Qualitative Comparisons.}
Fig.~\ref{fig:comparison_test} provides visualization results of the proposed \name{} on the COCO \textit{test-dev}.
Even with \textit{box-only} annotations, \name{} can output high-quality segmentation masks with fine boundaries.

\section{Conclusions}
In this paper, we explore the naive self-training with pseudo labeling for box-supervised instance segmentation, which is much limited by the noisy pseudo masks.
To address this issue, we present an effective training framework, namely \name{}, which contains a collaborative teacher and perturbed student for mutually generating high-quality masks and training with pseudo masks.
We adopt mask-aware confidence scores to measure the quality of pseudo masks and noise-aware mask loss to train the student with pseudo masks. 
In the experiments, \name{} achieves promising improvements on COCO, PASCAL VOC, and Cityscapes datasets, indicating that the proposed training framework is effective and can achieve higher level of weakly supervised instance segmentation.

% \paragraph{Acknowledgement.} This work was in part supported by NSFC (No. 61733007).

\appendix
\section{Data Augmentation}

We adopt two levels of data augmentation, \ie, strong augmentation and weak augmentation, and provide the augmentation details in Tab.~\ref{tab:augmentation_def}, which mainly perturb the image colors.
In addition, we adopt the multi-scale augmentation~\cite{TianSC20} (\ie, multi-scale training) and random flip for images fed into the student.

\begin{table}[hp]
    \centering
    \caption{\textbf{Specifications of Data Augmentation.}}
    \vspace{-5pt}
    \renewcommand{\tabcolsep}{7pt}
    \renewcommand{\arraystretch}{1.1}
    \small
    \begin{tabular}{c|c|c}
    Aug. & Prob. & Parameters \\
    \tline
    \multicolumn{3}{c}{Weak Augmentation} \\
    \hline
    Color Jittering & $0.2$ & \makecell[c]{brightness ($0.2$), contrast ($0.2$), \\saturation ($0.2$), hue ($0.1$) } \\
    \hline
    \multicolumn{3}{c}{Strong Augmentation} \\
    \tline
    Color Jittering & $0.8$ & \makecell[c]{brightness ($0.4$), contrast ($0.4$), \\saturation ($0.4$), hue ($0.1$)} \\
    \hline
    Grayscale & $0.2$ & - \\
    \hline
    Gaussian Blur & 0.5 & kernel = $3$, sigma = ($0.1$, $2.0$) \\
    \end{tabular}
    \label{tab:augmentation_def}
    \vspace{-5pt}
\end{table}

%%%%%%%%% REFERENCES
{\small
\bibliographystyle{ieee_fullname}
\bibliography{egbib}
}

\end{document}